\documentclass[a4paper, 10pt, conference]{ieeeconf}      

\IEEEoverridecommandlockouts                              

\usepackage{cite}
\usepackage{amsmath,amssymb,amsfonts}
\usepackage{algorithmic}
\usepackage{graphicx, overpic}
\usepackage{textcomp}
\usepackage{hyperref}
\usepackage{xcolor}
\usepackage{pifont}
\usepackage{multirow}   
\usepackage{booktabs}   
\usepackage[labelfont=bf]{caption}
\usepackage{hyperref}
\usepackage{subcaption}
\usepackage{array}
\usepackage[most]{tcolorbox}
\usepackage{verbatim}
\usepackage[export]{adjustbox} 
\usepackage{tabularx}
\usepackage{booktabs}

\tcbuselibrary{listings, skins, breakable}
\usepackage{threeparttable}

\def\BibTeX{{\rm B\kern-.05em{\sc i\kern-.025em b}\kern-.08em
    T\kern-.1667em\lower.7ex\hbox{E}\kern-.125emX}}
\begin{document}

\title{ROBOCYCLE: Autonomous Dual-Arm Robotic Manipulation and Coordination for Recycling Applications}

\author{%
Rubén de J. Hilario-Cruz $^{a}$, Jesus A. García-González $^{a}$,  Enrique Coronado$^{b}$ and Arturo E. Cerón-López$^{a}$
\thanks{$^{a}$ Tecnologico de Monterrey, School of Engineering and Sciences, Department of Computer Science, Ave. Eugenio Garza Sada 2501, Monterrey 64849, NL, Mexico.}
\thanks{$^{b}$ National Institute of Advanced Industrial Science and Technology (AIST) Tokyo Waterfront, 2 Chome-3-26 Aomi, Koto City, Tokyo 135-0064, Japan.}
}

\maketitle
\begingroup
\addtocounter{footnote}{-1}
\endgroup

\begin{abstract}
As urban waste volumes escalate and labor shortages intensify, automated waste sorting systems are becoming a necessity. However, current robotic solutions often struggle with the 3D perception and manipulation of transparent, deformable, or cluttered objects. This work introduces ROBOCYCLE, an autonomous dual-arm robotic recycling platform designed to meet the recycling standards of the Tokyo metropolitan area. Our approach integrates multi-view RGB-D perception, transformer-based instance segmentation using RF-DETR, and 6-DoF grasp planning via the Anygrasp SDK. By processing segmentated point clouds, the system generates robust candidate poses for irregular and deformable waste. The system achieved a 90.3\% grasp success rate and 84.3\% overall task success rate, effectively performing complex coordinated tasks such as unscrewing PET bottle caps. The proposed platform offers a scalable solution for autonomous waste management in real-world human environments.
\end{abstract}

\section{Introduction}

One of the major sustainability challenges facing both developing and developed countries in the transition toward Industry 5.0 \cite{mesjasz2023can, coronado2022evaluating} and Society 5.0 \cite{alimohammadlou2023role} is the proper collection, sorting, and disposal of solid waste. This issue is both highly relevant and urgent, as inadequate waste management can pose serious health hazards to humans, plants, and animals due to the potential release of toxic substances into water, soil, and air \cite{bhattstate, lakhouit2025revolutionizing}. In this context, it is estimated that approximately 2 billion tons of solid waste are generated annually worldwide, of which around 400 million tons consist of plastic. However, less than 10\% of this plastic waste is recycled \cite{kaza2018waste, lakhouit2025revolutionizing}. This issue is partly attributed to the fact that waste sorting is often perceived as a tedious and unpleasant task, resulting in incorrect waste disposal due to low engagement or limited awareness of waste management practices \cite{roy2022if,pedersen2020hidden}.

Recent advances in Artificial Intelligence (AI), robotics, and the Internet of Things (IoT) present significant opportunities to improve waste management on human-centric environments \cite{lakhouit2025revolutionizing}. While existing robotic waste-sorting systems have improved the efficiency of industrial pick-and-place operations, many rely on parallel robot architectures optimized for repetitive tasks in structured environments. However, more complex waste-processing activities involving coordinated actions, object reorientation, and sequential manipulation may require greater flexibility. As an initial step toward more flexible and human-like robotic waste-processing systems, this work proposes a robotic framework integrating coordinated dual-arm manipulation, multi-view object perception, and intelligent grasp planning within a ROS 2-based architecture.

\section{Related Work}

Despite its significance for environmental and human well-being, limited research has been conducted on automated garbage classification \cite{prova2024garbage}. Some of most recent efforts in this area are summarized in Table \ref{tab:hybrid_comparison} and \ref{tab:metrics_comparison}. In this context, Koskinopoulou et al. \cite{koskinopoulou2021robotic} developed a vision-based industrial sorting system for material recovery facilities, integrating Mask R-CNN \cite{hmidani2022comprehensive} with an ABB IRB360 delta robot. Similarly, Cheng et al. \cite{cheng2024optimizing} proposed an AI-powered system for beverage container recycling in Japan, utilizing YOLOv7-based \cite{hussain2023yolo} for object recognition paired with suction-based grasping. Posteriorly,  Duan \cite{duan2024intelligent} presented a waste classification pipeline using DenseNet121 \cite{setyono2018betawi} and a 7-DoF simulated robotic arm. Most recently, Karim et al. \cite{karim2025rtdrnet} introduced RTDRNet-lite, a lightweight RT-DETR-based model \cite{carion2020end} for waste sorting validated on a 4-DoF robotic arm. However, all these frameworks formulate waste sorting as a 2D planar pick-and-place problem, primarily relying on conveyor-based parallel or single-arm robots \cite{abdu2022survey, deng2025efficient}. 

The integration of humanoids, mobile manipulators, and service robots into domestic and public spaces presents new opportunities for more dexterous and flexible robotic applications. Specifically, the combination of dual-arm manipulation with 3D waste localization and grasping. As shown in Table \ref{tab:hybrid_comparison}, previous works are limited to simple pick-and-place tasks, without incorporating RGB-D perception or physically grounded grasp generation. As a result, complex tasks such as handling transparent PET bottles, removing caps and labels, or manipulating deformable objects are still challenging. Moreover, the shift from 2D methods to full 3D pose estimation and grasping, introduces additional difficulties, including sensitivity to lighting variations, detection of small or distant objects, inaccuracies in 3D localization, and errors caused by deformation, degradation, and occlusion in unstructured environments \cite{abdu2022survey}. 

\begin{table*}[t]
\centering
\caption{Comparison of architectural design and grasping scope of robotic waste sorting systems.}
\label{tab:hybrid_comparison}
\renewcommand{\arraystretch}{1.2}
\begin{tabular}{p{2.2cm} p{2.2cm} p{2.2cm} p{2.2cm} c c c}
\hline
\textbf{Work} 
& \textbf{Architecture} 
& \textbf{Perception} 
& \textbf{Grasp Strategy} 
& \textbf{Dual-arm} 
& \textbf{Transparent Obj.} 
& \textbf{Deformable Obj.} \\
\hline

Koskinopoulou (2021) \cite{koskinopoulou2021robotic}
& Delta robot (ABB IRB360) 
& Mask R-CNN (stereo) 
& Planar suction PnP 
& \ding{55} & \ding{55} & \ding{55} \\

Cheng (2024) \cite{cheng2024optimizing} 
& Delta robot (HIWIN) 
& YOLOv7 + SAM 
& 2D pick-point (suction) 
& \ding{55} & $\triangle$ & $\triangle$ \\

Duan (2024) \cite{duan2024intelligent}
& Simulated 7-DoF arm 
& DenseNet121 (classification) 
& Scripted (simulation) 
& \ding{55} & \ding{55} & \ding{55} \\

Karim (2025) \cite{karim2025rtdrnet}
& 4-DoF robotic arm 
& RTDRNet-lite (RT-DETR) 
& 2D IK-based grasp 
& \ding{55} & \ding{55} & \ding{55} \\

\textbf{ROBOCYCLE (Ours)}
& Dual-arm (UR5e + xArm7) 
& RGB-D + RF-DETR 
& 6-DoF grasp generation
& \ding{51} & \ding{51} & \ding{51} \\

\hline
\end{tabular}

\vspace{1mm}
\footnotesize{
\ding{51}: supported \quad \ding{55}: not supported \quad $\triangle$: partially supported \\
}
\end{table*}

\begin{table*}[t]
\centering
\footnotesize
\caption{Comparison of detection capabilities and evaluation environments of robotic waste sorting systems.}
\label{tab:metrics_comparison}
\renewcommand{\arraystretch}{1.2}

\begin{tabular}{
p{2.5cm}
p{2.3cm}
p{2.2cm}
p{1.8cm}
p{1.5cm}
p{2.4cm}
p{2.3cm}
}
\toprule

\textbf{Work} &
\textbf{Vision Task} &
\textbf{Objects} &
\textbf{Accuracy (\%)} &
\textbf{RGB-D} &
\textbf{Complex Manip.} &
\textbf{Evaluation} \\

\midrule

Koskinopoulou (2021) \cite{koskinopoulou2021robotic} &
Detection + classification &
Industrial recyclables &
$\sim$91.8 &
Stereo &
No &
Industrial facility \\

Cheng (2024) \cite{cheng2024optimizing} &
Detection + classification &
Beverage containers &
93 &
RGB &
Limited &
Real deployment \\

Duan (2024) \cite{duan2024intelligent} &
Image classification &
12 waste classes &
$>$90 (val.) &
RGB &
No &
Simulation \\

Karim (2025) \cite{karim2025rtdrnet} &
Real-time detection &
4 material classes &
-- &
RGB &
No &
Real hardware \\

\textbf{ROBOCYCLE (Ours)} &
Detection + classification &
Mixed household recyclables &
97.8 &
Multi-view RGB-D &
Yes (unscrewing, processing) &
Real hardware \\

\bottomrule
\end{tabular}
\end{table*}

\section{Contributions}

The primary contribution of this work is the development of ROBOCYCLE, a dual-arm robotic recycling architecture designed to handle complex waste-sorting tasks on human-centric environments. For this, we integrated a multi-view RGB-D perception pipeline that leverages RF-DETR-based detection and segmentation, combined with point-cloud reconstruction, to achieve precise 3D localization of recyclable waste under unstructured conditions. The proposed approach allows to address scenarios requiring interaction beyond simple object transfer, such as removing caps from bottles under realistic conditions. By combining transformer-based perception with 6-DoF grasping, the proposed system enables complex dual-arm coordination tasks aligned with Japanese PET recycling standards and global Sustainable Development Goals (SDGs), particularly SDG 12 and SDG 13 \cite{chauhan2026revolutionizing}. This positions ROBOCYCLE as a scalable, human-centered solution for modern circular economies.

\section{System Architecture and Methodology}

\begin{figure*}[htbp]
    \centering
    \includegraphics[width=0.85\linewidth]{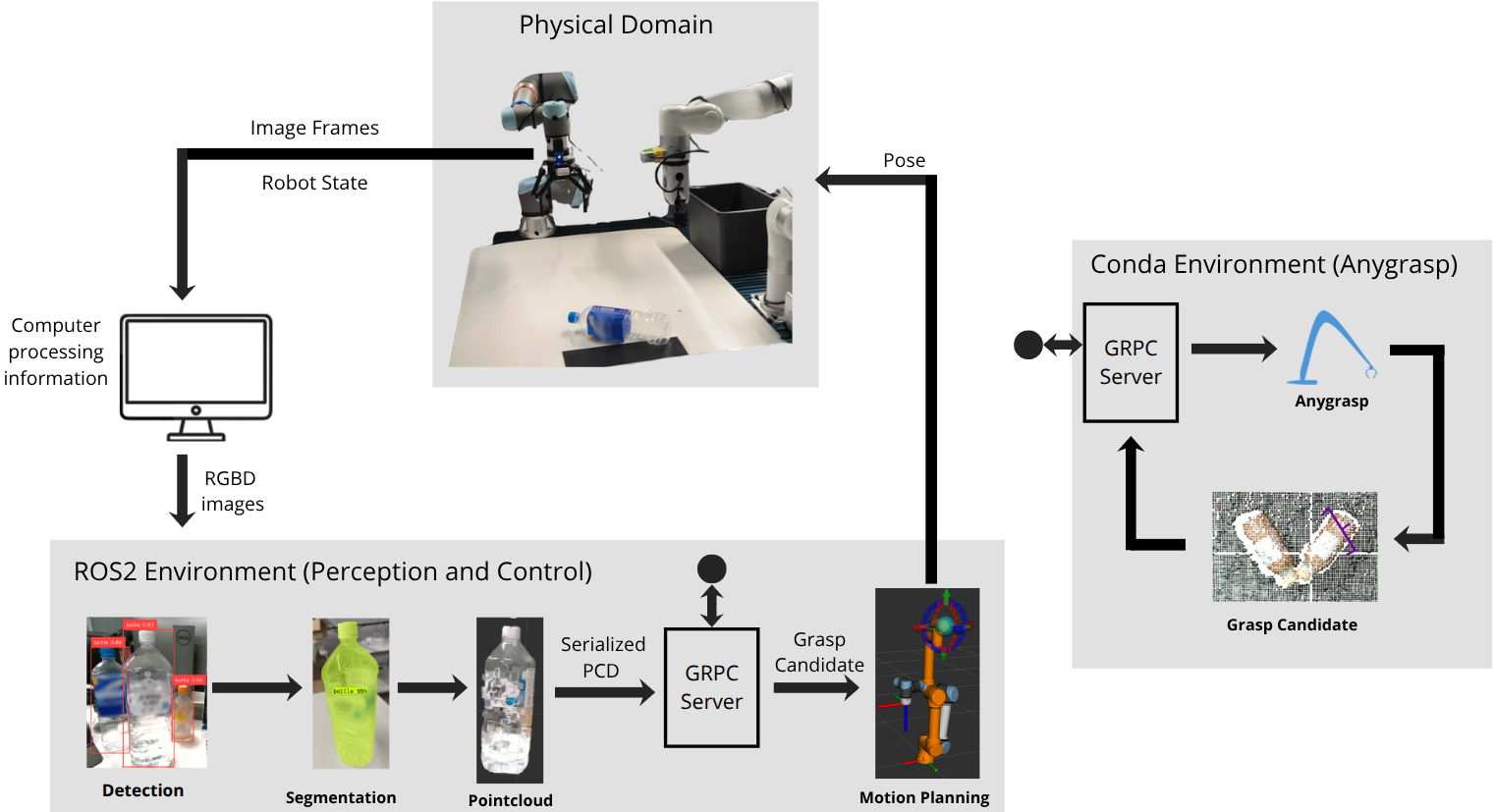}
     \caption{System overview showing the integration of key components for robot perception and grasp planning.}
    \label{fig:overview}
\end{figure*}

\begin{figure*}
    \centering
    \includegraphics[width=0.88\linewidth]{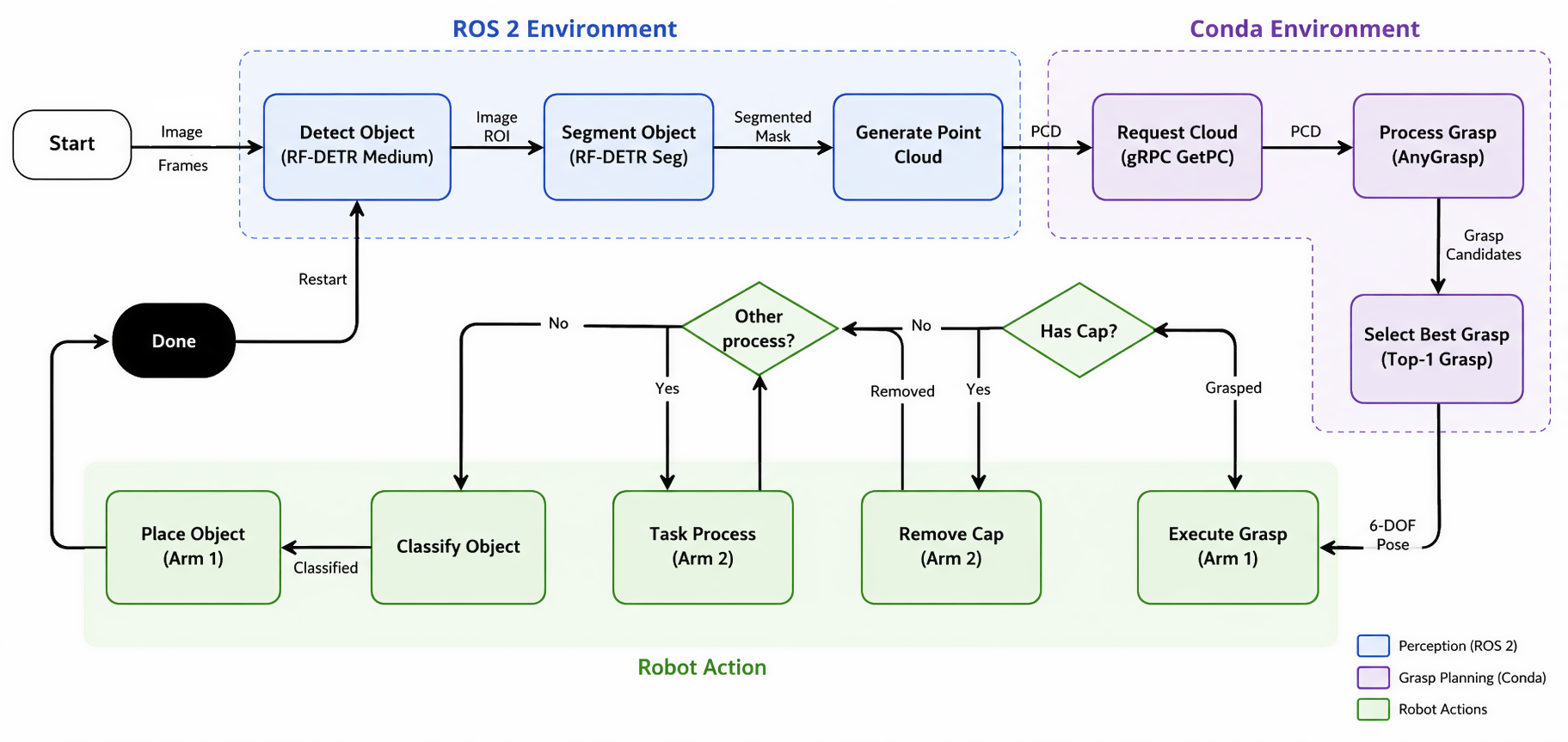}
     \caption{Flux diagram of the workflow for the robot's perception and grasp planning system.}
     \label{fig:flux_diagram}
\end{figure*}

Figure~\ref{fig:overview} illustrates a general overview of the proposed system and the relationships between its main modules.  The physical domain consists of a Universal Robots UR5e and a UFactory xArm7 mounted on opposite sides of a shared workspace. Both manipulators are equipped with Robotiq grippers and controlled through ROS~2 drivers integrated with MoveIt~2. Two RealSense D405 RGB-D cameras are mounted in an eye-in-hand configuration, providing high-resolution depth images at close range. The transformation between the D405 camera frame and robot base frame is estimated using a calibration board and solved via hand–eye calibration. This ensured that 3D points extracted from segmentation masks align correctly with the robot’s workspace. The cameras publish synchronized RGB and depth frames through \textit{the topics '/camera/camera/color/image\_rect\_raw' }, \textit{'/camera/camera/depth/image\_raw'} and \textit{'/camera/camera/camera\_info'}. The perception and planning modules run on a dedicated workstation equipped with an NVIDIA RTX 3080. The robotic arms are controlled and synchronized through ROS 2 libraries, using behavior trees \cite{coronado2019robots} for coordinating their execution. Figure~\ref{fig:flux_diagram} shows the proposed workflow for dual-arm manipulation.  The architecture is distributed across a ROS 2 environment for data collection and a Conda-based AI environment integrating AnyGrasp used to avoid compatibility issues. Additionally, a robot action module is used to manage the integrated logic for waste classification. The functionality of each module is briefly described below.

\subsection{ROS 2 environment for object detection}

Two main frameworks were integrated and compared for object detection in the RO2 environment: YOLOv11 and RF-DETR. A detailed comparison of these approaches is presented in section \ref{se:comp}. Using camera intrinsics and depth images, the 2D segmentation mask is projected into 3D to obtain a dense point cloud for each object. Then a denoising and downsampling stage is applied to remove isolated points and reduce computational load. The resulting point cloud is then transformed into the robot base coordinate frame using precomputed extrinsic calibration between the D405 sensor and the robot workspace. Finally, the point cloud is published on the \texttt{/camera\_inference/point\_cloud}. 

\subsection{Conda AI environment for grasp planning}

Prior to adopting AnyGrasp for 6-DoF grasp synthesis, a more traditional pose estimation approach was implemented using point-cloud-based feature matching and geometric alignment.  Although this method produced reasonable pose estimates for rigid and opaque objects, its performance degraded rapidly for transparent PET, deformable items, and objects with weak geometric features. This motivated the transition to AnyGrasp, which is able to generate physically plausible grasp candidates regardless of object symmetry, texture, or transparency. 

\begin{figure}[h]
    \centering
    \includegraphics[width=0.9\linewidth]{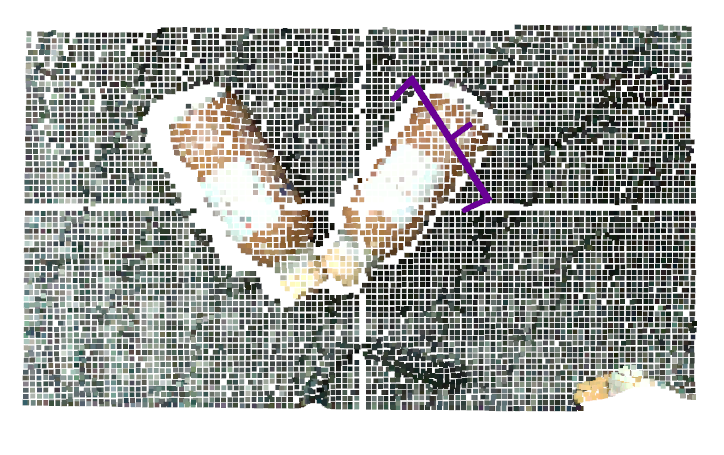}
    \caption{3D point cloud of an object with the best 6-DoF grasp candidate overlaid. The grasp pose, shown in purple, represents the optimal position and orientation for the robot's end effector to securely grasp the object, as determined by the AnyGrasp SDK. The point cloud visualizes the object's surface geometry.}
    \label{fig:anygrasp}
\end{figure}

In our approach, the segmented and filtered point cloud is provided to the AnyGrasp SDK, which estimates physically plausible and kinematically feasible 6-DoF grasp poses. Using surface geometry and contact stability analysis, AnyGrasp generates a ranked set of grasp candidates. The grasp with the highest predicted success probability is then selected for each detected object. An example of the AnyGrasp output is shown in Figure~\ref{fig:anygrasp}.

\begin{figure}[h]
    \centering
    \includegraphics[width=0.3\linewidth]{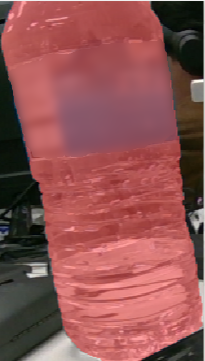}
    \includegraphics[width=0.5\linewidth]{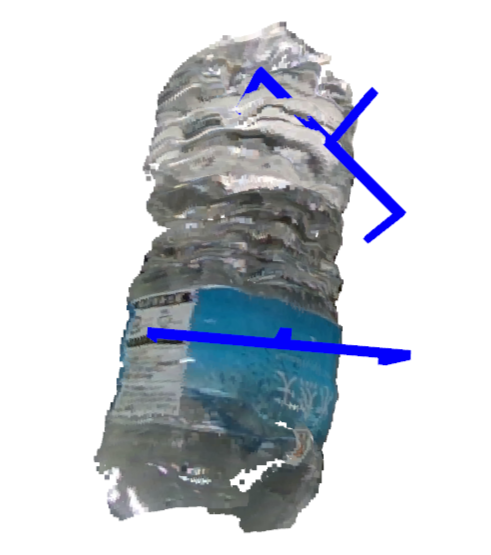}
    \caption{
    Left image shows the object segementation mask in RGB image obtained from the camera topics.
    Right image shows the cloud point obtained from the segmentation mask and the best grasp candidate.}
    \label{fig:mask_to_grasp}
  \end{figure}

Figure ~\ref{fig:mask_to_grasp} shows the object mask in RGB image and its final grasp candidate elected after processing the point cloud with Anygrasp SDK.

A gRPC server was integrated to enable communication between ROS2 modules and the Conda-based AI environment. In particular, this server allows a client to request the point cloud and receive the segmented cloud with the XYZ coordinates and the colors of each point, extracted from the depth and segmentation images. For this, the point cloud is subscribed to the \texttt{/camera\_inference/point\_cloud} topic and stored in the \texttt{latest\_points} variable. The server also allows a client to send the calculated grasp pose, which is then published on the \texttt{/grasp\_pose} topic, which enable the robot to execute the grasp. 

\subsection{Grasp Selection and Feasibility Checking}

Before execution, each candidate pose is validated against: i)  kinematic reachability of both arms; ii) potential inter-arm and environment collisions, iii) grasp approach direction relative to the table plane and iv) the recycling class assigned to the item. Then, the system assigns each object to either the UR5e or xArm7 based on each robot's proximity to the target and the spatial constraints required for collision avoidance.

\subsection{Motion Planning and Dual-Arm Coordination}

We used MoveIt~2 to generate collision-free trajectories using the RRT-Connect and Pilz planners. Grasp poses from AnyGrasp are converted into target frames, and the system computes inverse kinematics solutions that account for joint limits, workspace sharing, and inter-arm constraints.

Trajectory Execution is handled by ROS~2 controllers (\textit{`joint\_trajectory\_controller`}) running on each robot. A temporal synchronization node ensures that: i) Only one arm enters the grasping zone at a time; ii) Both arms can operate in parallel when removing objects from the workspace and iii) Error recovery behaviors are consistent across both robots.

\subsection{Recycling Classification and Object Sorting}

Each detected object is assigned a category based on RF-DETR classification and subsequently routed to the corresponding disposal bin (e.g., plastic, PET, metal, cap). After grasping, the robot follows a predefined drop-off trajectory to place the item in the correct container. The dual-arm configuration allows simultaneous sorting of multiple items, reducing cycle time and increasing throughput.
\begin{figure}[h]
    \centering
    \includegraphics[width=0.35\linewidth]{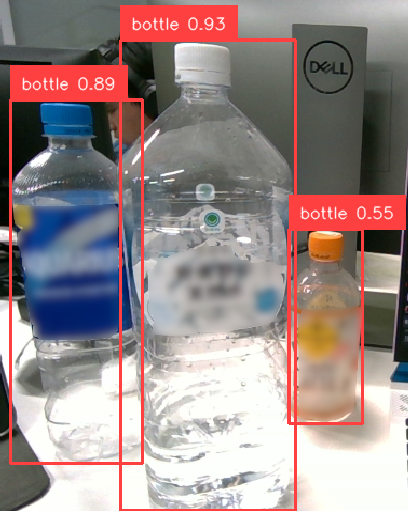}
    \includegraphics[width=0.55\linewidth]{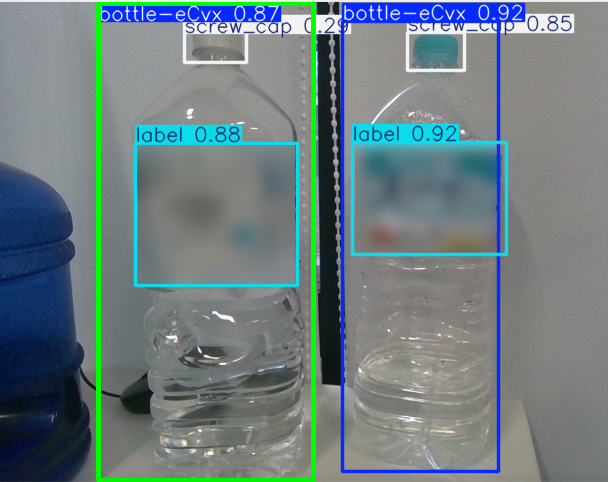} 
    \caption{RF-DETR outputs detecting and labeling multiple objects within the scene, and drawing bounding boxes around each identified object}
    \label{fig:image2}
  \end{figure}

\section{Experimental Setting and Results}
In this section, we report the performance of the proposed ROBOCYCLE system, focusing on the perception pipeline accuracy and the grasping success rate.

\subsection{Vision System Performance}
\label{se:comp}
We observed that YOLOv11 provided acceptable performance for coarse classification. However,  its segmentation masks were inconsistent in frames with very high lighting and its detection accuracy degraded significantly for transparent PET bottles and thin objects such as labels.
Compared to YOLOv11, RF-DETR achieved higher overall accuracy across all recyclable categories while maintaining a similar inference speed on the used workstation. These improvements were particularly evident in transparency, reflective materials and irregular bottles shapes. Given its robustness to occlusions, RF-DETR was selected as the primary detection and segmentation approach for the system. 

\begin{figure}[h]
    \centering
    \includegraphics[width=0.95\linewidth]{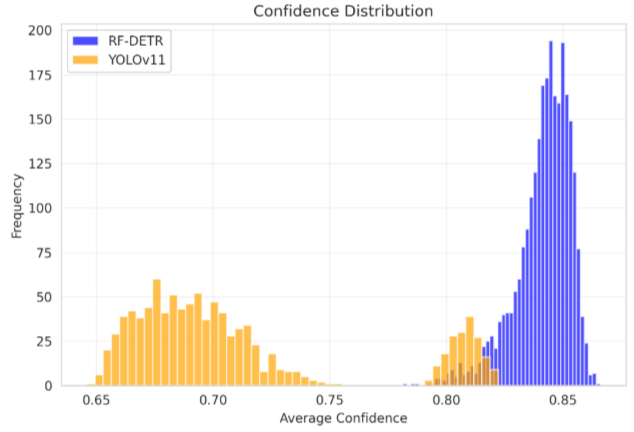}
    \caption{Confidence distribution comparison between the two vision models, YOLOv11(yellow) and RF-DETR(blue).}
    \label{fig:avg_precision}
\end{figure}

\begin{figure}[h]
    \centering
    \includegraphics[width=0.9\linewidth]{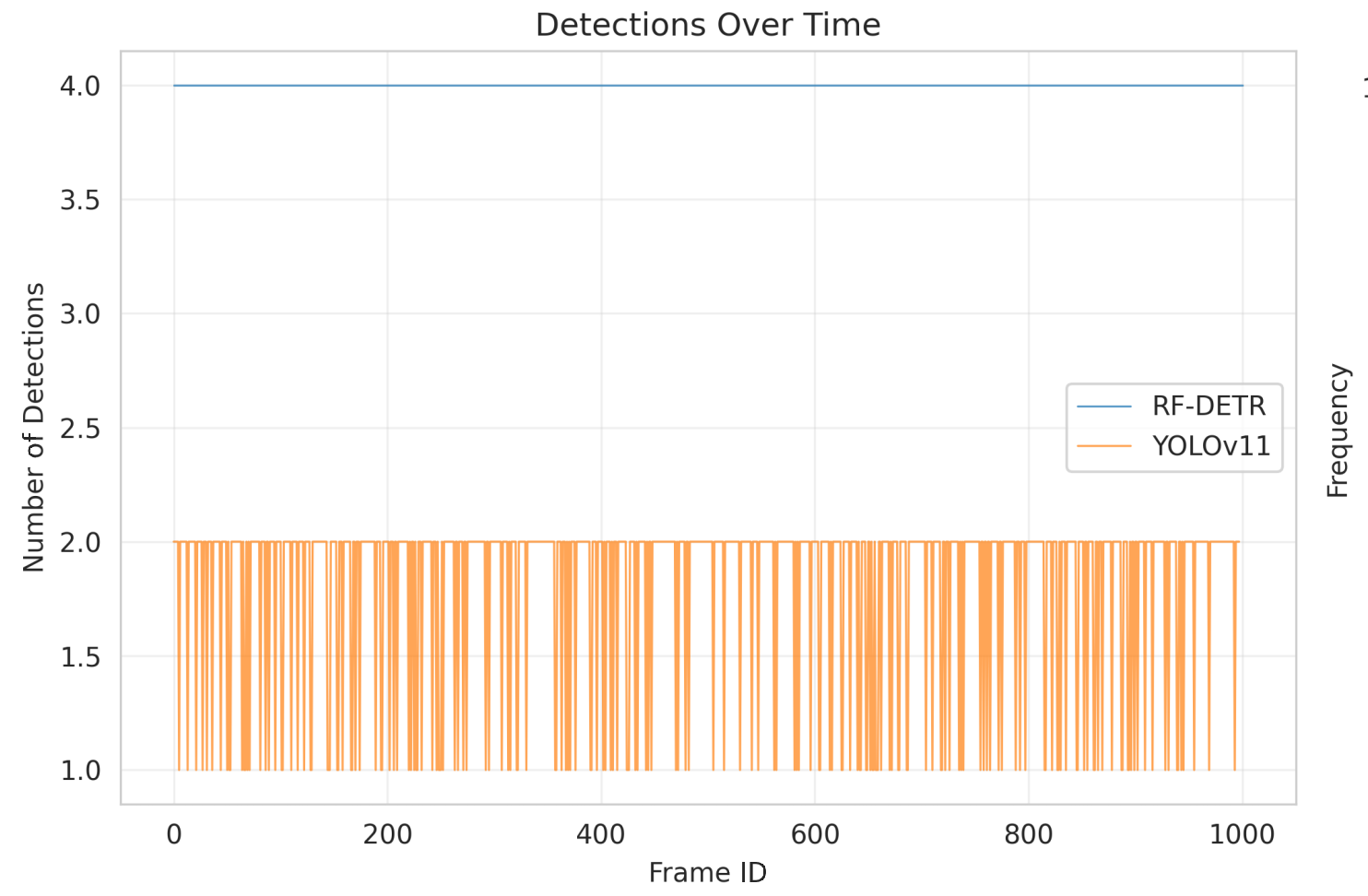}
    \caption{Comparison between the number of detections over time. RF-DETR maintains a constant count of 4.0 detections per frame, while YOLOv11 exhibits significant "flicker," fluctuating inconsistently between 1.0 and 2.0 detections.}
    \label{fig:d_time}
\end{figure}

Figures \ref{fig:avg_precision} and \ref{fig:d_time} illustrate the comparative performance between both models. Our results indicate that RF-DETR achieved a mean Average Precision (mAP) of 0.8416, significantly outperforming the 0.71 mAP of the YOLOv11 model. This performance was validated through live camera streams, where the model sustained real-time inference during active robot manipulation. Unlike previous iterations, which caused inconsistencies, the model demonstrated superior robustness in detecting transparent PET bottles and thin labels under dynamic lighting and motion.

Furthermore, the inference time was measured to ensure real-time viability. The system achieved an average processing speed of 20.45 FPS on an NVIDIA RTX 3080, which is sufficient for the dynamic sorting tasks required in the ROBOCYCLE platform. Figure \ref{fig:d_time} shows a clear difference in detection stability. RF-DETR (blue) maintains a perfectly flat line at 4.0 detections, indicating it consistently identifies the same number of objects in every frame without flickering. Conversely, YOLOv11 (orange) exhibits high instability, constantly fluctuating between 1 and 2 detections, which suggests frequent "dropped" objects or false negatives throughout the sequence.

\subsection{Grasping and manipulation}
To evaluate the AnyGrasp integration, we conducted a series of 100 physical trials with different object distributions across the robots' workspace. We obtained a grasping success rate of 90.3\%. The main failure modes, were attributed to sensor noise in highly reflective surfaces. However, the detection pipeline significantly reduced grasping failures by avoiding errors typical of deformable objects, a significant advantage over traditional 2D top-down grasping methods.

\subsection{Full System Evaluation}
To evaluate the suitability of the fully integrated ROBOCYCLE architecture, a set of sorting and processing tasks were performed. This experiment tested the entire pipeline: from initial RGB-D capture and RF-DETR segmentation to the execution of specialized manipulation primitives, such as the unscrewing of PET bottle caps and label removal. Overall, the integrated system achieved a task success rate of 84.3\%. The results for each stage of the complete cycle are summarized as follows:

\subsubsection{Object Classification and Localization}  The system correctly identified and localized items with a precision of 97.8\%, providing stable point clouds for the subsequent stages.

\subsubsection{Complex Manipulation (Unscrewing/Processing)}  For PET bottles and certain types of cans, the coordinated movement between the xArm7 (holding the body) and the UR5e (executing the rotation of the cap) achieved a success rate of 84.3\%. Failures in this stage were primarily due to slight misalignments caused by the initial grasp, which generated a small inclination in the objects.

\subsubsection{System Throughput:} The complete process, including classification, pick-and-place, and specialized processing (unscrewing), resulted in an average cycle time of 8.3 seconds per item, and 15.6 seconds when processing com plex manipulation tasks

The experimental results presented in this section demonstrate that dual-arm coordination provides the necessary geometric dexterity and operational redundancy required to effectively process the recycling of complex waste.

\begin{figure}[h]
    \centering
    \includegraphics[width=0.7\linewidth]{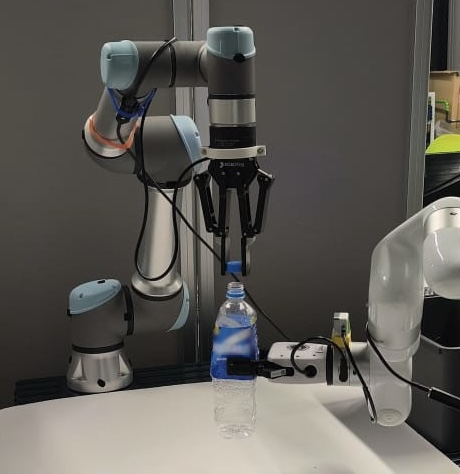}
    \caption{Figure representing complex task manipulation tasks such as removing caps from bottles.}
    \label{fig:complex_man}
\end{figure}

\section{Conclusion}
We introduced ROBOCYCLE, an integrated dual-arm robotic platform designed to automate the recycling process through coordinated manipulation. The proposed system integrated multi-view perception, deep learning–based segmentation and object detection, point cloud extraction, and 6-DoF grasp generation with coordinated dual-arm manipulation. This unified pipeline enabled the automated sorting of common solid waste often generated in homes and office environments, such as plastic bottles and aluminum cans, in accordance with the recycling guidelines of the Tokyo metropolitan area. We expect that the proposed system can inspire further research toward the development of feasible and scalable solutions for autonomous waste management, capable of effectively addressing the challenges typically encountered in real-world (i.e., non-controlled) and human environments. 

While the current implementation of ROBOCYCLE focuses on real-time perception and physically grounded dual-arm manipulation, recent advances in Vision--Language and Vision--Language--Action (VLM/VLA) models open promising avenues for extending the system toward higher-level autonomy. These models have demonstrated strong capabilities in semantic reasoning, task planning, and instruction following, which are complementary to the low-level execution strengths of the proposed platform.

Rather than replacing the deterministic perception and grasping pipeline presented in this work, VLM/VLA models could operate as a high-level decision-making layer, enabling semantic interpretation of waste objects, reasoning over recycling rules, and dynamic task selection based on natural language instructions or environmental context. In this sense, ROBOCYCLE can be viewed as a real-time manipulation backbone that remains compatible with emerging cognitive models, allowing future upgrades without redesigning the core robotic architecture.

Furthermore, the proposed manipulation framework is not limited to static workcells. A natural extension of this research is the integration of the ROBOCYCLE manipulation stack with a mobile platform such as the TIAGo Pro robot. Combining reliable 6-DoF grasping and dual-arm coordination with mobile navigation would enable complex recycling tasks in human-centered environments, including offices, campuses, and public facilities.

\bibliographystyle{IEEEtran}
\bibliography{referenceetal}

\end{document}